\documentclass[12pt,a4paper]{article}
\usepackage{epsf,exscale,times}
\usepackage[english]{babel}
\usepackage{graphicx}
\usepackage{amsmath}
\usepackage{fancyhdr}
\usepackage{a4wide}

\usepackage[T1]{fontenc}
\usepackage{graphicx}
\usepackage{hyperref}
\usepackage{longtable}
\usepackage{array}
\usepackage{tikz}
\usepackage{booktabs}
\usepackage{subcaption}
\usepackage{amsmath}
\usepackage{amssymb}
\usepackage{cleveref}
\usepackage{dsfont}
\usepackage{color}
\hypersetup{colorlinks,
linkcolor={blue},
citecolor={blue},
urlcolor={blue}}

\newcommand{\email}[1]{\href{mailto:#1}{\texttt{#1}}}
\title{Tracking mulitple targets with multiple radars using Distributed Auctions\footnotetext{Published in: \textit{The International Radar Symposium IRS 2023, May 24-26, 2023, Berlin}}}
		\author{Pierre Larrenie\thanks{
\textit{LIGM}\newline
 	\textit{Univ. Gustave Eiffel, CNRS, Marne-la-Vallée, France}\newline
 	\email{pierre.larrenie@esiee.fr}\newline THALES SIX GTS France\newline
		4 Av. des Louvresses, 92230, Gennevilliers\newline
		France\newline
		\email{pierre.larrenie@thalesgroup.com}}
	\and Cédric L R Buron\thanks{KlaIM, L@bISEN\newline
		33Q avenue du champ de Manœuvre 44470 Carquefou\newline
		France\newline
		\email{cedric.buron@isen-ouest.yncrea.fr}} \and
	Frédéric Barbaresco\thanks{THALES LAND \& AIR SYSTEMS\newline
		6 rue de la Verrerie, 92190 Meudon,\newline
		France\newline\email{frederic.barbaresco@thalesgroup.com}}}

\date{}
\begin{document}

\maketitle

\renewcommand{\headrulewidth}{0pt}
\renewcommand{\footrulewidth}{1pt}
\renewcommand{\footskip}{50pt}

\pagestyle{plain}

%%----------------------------------------------------
%% Insert the title of your paper between the brackets
%%----------------------------------------------------
%
% \IEEEauthorblockA{\IEEEauthorrefmark{1} 
% 	}
% \IEEEauthorblockA{\IEEEauthorrefmark{2} \textit{LIGM}\\
% 	\textit{Univ. Gustave Eiffel, CNRS, Marne-la-Vallée, France}\\
% 	\email{pierre.larrenie@esiee.fr}\\
% 	}
% \IEEEauthorblockA{\textit{KlaIM}, \\
%         \textit{L@bISEN, ISEN Ouest, Carquefou, France}\\\email{cedric.buron@isen-ouest.yncrea.fr}\\}
% \IEEEauthorblockA{}
% }
% \author{Pierre Larrenie\inst{1} \and
% Cédric L R Buron\inst{2} \and
% Frédéric Barbaresco\inst{3}}
%
% \authorrunning{P. Larrenie et al.}
% First names are abbreviated in the running head.
% If there are more than two authors, 'et al.' is used.
% %
% \institute{
% 	THALES SIX GTS France\\
% 	4 Av. des Louvresses, 92230, Gennevilliers\\
% 	France\\
% 	\email{pierre.larrenie@thalesgroup.com}
% 	\and
% 	KlaIM, L@bISEN\\
% 	33Q avenue du champ de Manœuvre 44470 Carquefou\\
% 	France\\
% 	\email{cedric.buron@isen-ouest.yncrea.fr}
% 	\and
% 	THALES LAND \& AIR SYSTEMS\\
% 	6 rue de la Verrerie, 92190 Meudon,\\
% 	France\\
% 	\email{frederic.barbaresco@thalesgroup.com}}
% %              % typeset the header of the contribution
%
\abstract{
Coordination of radars can be performed in various ways. To be more resilient
radar networks can be coordinated in a decentralized way. In this paper,
we introduce a highly resilient algorithm for radar coordination based
on decentralized and collaborative bundle auctions. We
first formalize our problem as a constrained optimization problem
and apply a market-based algorithm to provide an approximate solution. Our
approach allows to track simultaneously multiple targets, and to use
up to two radars tracking the same target to improve accuracy. We
show that our approach performs sensibly as well as a centralized approach
relying on a MIP solver, and depending on the situations, may outperform it
or be outperformed.\\
% In this paper, we present an algorithm allocating tasks for a set of static autonomous radars with
% rotating antennas. It allows a set of radars to allocate in a fully
% decentralized way a set of tasks for tracking targets according to their
% location, considering that tracking a target with several can improve accuracy. The
% allocation algorithm proceeds through a collaborative and fully
% decentralized auction protocol, using a collaborative auction protocol
% (Consensus Based Bundle Auction algorithm). Our algorithm is based on a
% double use of our allocation protocol among the radars. The latter begin
% by allocating targets, then launch a second round of allocation if they
% have resources left, in order to improve accuracy on targets already
% tracked.\\
\textbf{Collaborative combat, Distributed Auctions, Multi-Radar	Tracker}
}

\section{Introduction}\label{presentation-of-the-problem-of-collaborative-multi-radar-tracking}
·The management of small and cheap sensors is a rising topic, both for civil and defence applications.
In particular, recent advances have been made in the management of small radars, both in a centralized
and a decentralized way. Decentralized approaches provide the system with more robust coordination,
but are often expected to perform more poorly than the centralized approaches. Still, some methods have
shown to be very effective, and are indeed widely used in robotics, such as auction-based ones.

In this paper, we deal with the problem of target allocation among a set of radars with
rotating antennas. In order to have a robust approach, we propose a fully decentralized
approach. Our approach allows to allocate each target to one or two radars, getting lower
uncertainty on the position of the target when two radars track them. Our method relies on
collaborative auction. This method is among the ones that are
fast enough to cope with the robotics issues \cite{gini2017multi}. To cope
with the specificity of our problem --the possibility to reduce the uncertainty on the
location of the target-- we propose to use 2 turns of the same allocation algorithm.
The first one ensures as many targets as possible are tracked, while the second one
allocates remaining resources to already-tracked targets in order to reduce uncertainty.

In order to take dynamism, \emph{i.e.} moving targets in account,
the algorithm is restarted on a regular basis.
It ensures that a bid made by a radar can decrease when the target moves
away from it.
In our applications, the targets move quickly, and their movement
cannot be predicted on the long term, we therefore rely on a collaborative auctions
algorithm that does not plan the following rounds.
Our algorithm is however based on radars
capable of anticipating the short-term positions of targets, thanks to a
Kalman filter. We test our approach on multi-radar
tracking scenarios where the radars, autonomous, must follow a set of
targets in order to reduce the position uncertainty of the targets.
Surveillance aspects are not taken in account in this paper. It is assumed
that the radars can pick up targets in active tracking, with an area of
uncertainty corresponding to their distance.

The paper is organized as follows: we first introduce existing works for multi-radar
multi-target allocation. Section~3 presents the problem, with the
radar model we use, and the mathematical optimization problem formalizing the
target allocation for a set of radars. We then present our approach based on
a double consensus-based bundle auction, and the simulation we have been using for
the evaluation of our method. Section~5 presents the results of our method
and compares it with a centralized approach relying on operational research.
We summarize our work and provide possible perspectives in section~6.

\hypertarget{related-works}{%
\section{Related works}\label{related-works}}

Several works have focused on the use of decentralized approaches for
task allocation for sensor, since the seminal work of Lesser et at.
\cite{lesser2003distributed}. Since then, many approaches have been used
among which Max-Sum based algorithms, market based algorithms, and
reactive approaches \cite{gini2017multi}. Recently, for real
time use cases, auction methods have gained much interest in the
multi-agent community \cite{krainin2007application} for their capacity to perform good
allocation in an affordable time. Many methods have been using auctions
since then to allocate tasks in real-time, including to robots, sensors
and radars (see for instance \cite{deliang2017distributed}).

One of the most successful recent algorithms is the Consensus-Based Bundle Auction algorithm (CBBA) \cite{choi2009consensus}, a decentralized auction-based allocation algorithm. This
algorithm allows to perform the allocation in a fully decentralized way,
the agents acting both as auctioneers and bidders. This algorithm has
since been used several times for sensors \cite{jia2017consensus,sameera2012robust}. However, none of
them has been taking into account the specificities of radars,
\emph{i.e.,} their collaboration through the intersection of their
uncertainty ellipses. Similarly, the challenge of high dynamicity has
been barely studied.

\section{Problem statement}
\label{problem}
In this section, we introduce the different elements of the problem. We first describe the model that has been used for the radars. We then introduce a definition of the multi-target multi-radar allocation problem in the formalism of Constraint Optimization Problems (COP).

\subsection{Radar model}
We consider that each radar has a 2-dimensional frame in a polar
coordinate system centered on itself. It is estimated that the influence
of elevation is negligible, so it is not useful to use a 3-dimensional
landmark.

Each target therefore has a position in the radar reference frame
determined by its distance, denoted \(r\) , and its azimuth (polar angle),
denoted \(\theta\).
The precision of the measurement made by the radar is noted:
\(\sigma_{r}\)  in distance and \(\sigma_{\theta}\)  in azimuth. The
resulting measurement uncertainty is represented as an ellipse. The
measure itself corresponds to a centered 2-dimensional Gaussian random variable of
covariance \mbox{$K = R(\theta)\begin{pmatrix}
\sigma_{r} & 0 \\
0 & \sigma_{\theta} \\
\end{pmatrix}$} , where \(R\left( \theta \right)\) represents the rotation matrix of angle \(\theta\). During active tracking, the aim is therefore
to anticipate the next measure on the target given its past
positions and its current position thanks to the use of a
Kalman filter. As for the measurement, there is also a prediction
uncertainty.
% This is summarized in \cref{fig-uncertainty-ellipses} with the measurement
% uncertainty in orange and the prediction uncertainty associated with the
% tracking of the target by the radar in purple.

The signal received by the radar is assumed to be subject to Gaussian
white noise. In order to reduce the size of the problem anbd keep it tractable, the signal to noise ratio (S/N)\footnote{The S/N (Signal to
Noise Ratio) corresponds to the ratio between the noise power of the
useful signal and the power of ambient noise.} is assumed to be constant. The value is set
to 13, which corresponds to a common value in practice. The S/N will
influence the standard deviation of the measures. The S/N 
corresponds to the quality of the output desired by the user. For a
given S/N it is possible to choose the parameters such as the wavelength
to be transmitted or the transmission power, etc.

\hypertarget{second-problem-multi-sensor-allocation}{%
\subsection{Multi-sensor multi-target allocation problem}\label{second-problem-multi-sensor-allocation}}

The problem of optimal multi-radar and multi-target allocation can be handled
in various ways. If we want to take account of the anticipation of future actions,
we can model it through Decentralized Partially Observable Markov Decision Processes. However, this problem is in the EXPTIME class (with a complexity in $O(\exp(\exp(n))$ where $n$ is the number of tasks, which is clearly not applicable to common application cases, and
even heuristic methods can take a very long time. Moreover, the target we consider
are highly mobile and their unpredictability makes the anticipation of the future very
hard to compute. We therefore propose to formalize the problem of optimal allocation on
each time step, not taking the future evolution in account. This problem can be formalized
as a constraint optimization problem:

\begin{equation*}
\begin{aligned}
\max\sum_{i,j,k}{c_{{ikj}} \cdot w_{{ikj}}} \\
\text{s.t.:} \\\begin{cases}
w_{{ikj}} = {x_{M}}_{{ij}} \land x_{O_{{kj}}},\forall\left( i,k \right) \in \mathcal{I}^{2}, \forall j\in \mathcal{J} & (A_{{ikj}}) \\
\sum_{i}w_{{ij}} \leq 1, \forall j\mathcal{\in J} & (C2) \\
\sum_{j}\gamma_{{ij}}\cdot ({x_{M}}_{{ij}} + x_{O_{{kj}}} - w_{{iij}}) \leq L_{t_{i}}, \forall i\in \mathcal{I} & (L) \\
\left( {x_{M}}_{{ij}},x_{O_{{kj}}} \right) \in \left\{ 0,1 \right\}^{2}, \forall\left( i,k,j \right)\mathcal{\in I \times I \times J} \\
w_{{ikj}} \in \left\{ 0,1 \right\},\forall\left( i,k \right) \in \mathcal{I}^{2}, \forall j\mathcal{\in J} \\
\end{cases}
\end{aligned} 
\end{equation*}
% \begin{equation*}
% \left( P2 \right):\left\{ \begin{aligned}
% \max\sum_{i,j,k}{c_{{ikj}} \cdot w_{{ikj}}} \\
% \text{s.t.:} \\
% w_{{ikj}} = {x_{M}}_{{ij}} \land x_{O_{{kj}}},\forall\left( i,k \right) \in \mathcal{I}^{2}, \\\forall j\in \mathcal{J}  (A_{{ikj}}) \\
% \sum_{i}w_{{ij}} \leq 1, \forall j\mathcal{\in J}  (C2) \\
% \sum_{j}\gamma_{{ij}}\cdot ({x_{M}}_{{ij}} + x_{O_{{kj}}} - w_{{iij}}) \leq L_{t_{i}}, \\\forall i\in \mathcal{I} (L) \\
% \left( {x_{M}}_{{ij}},x_{O_{{kj}}} \right) \in \left\{ 0,1 \right\}^{2}, \forall\left( i,j \right)\mathcal{\in I \times J} \\
% w_{{ikj}} \in \left\{ 0,1 \right\},\forall\left( i,k \right) \in \mathcal{I}^{2}, \forall j\mathcal{\in J} \\
% \end{aligned} \right.
% \end{equation*}
where
\begin{itemize}

\item \(\mathcal{I}\) is the set of radars and \(\mathcal{J}\) the set of tasks. Note that we are placed here, in the framework
\(\mathcal{I} \ll \mathcal{J}\).
\item
\(c_{ikj}\) : Corresponds to the utility that the radar \(i\) 
and the radar \(k\)  provide to the system if the radar \(i\)  handles
the task \(j\)  as a main radar and \(k\)  as an optional radar.
\(c_{ikj}\)  is of the following form, with
\(V(\mathcal{E}_{ij})\) (respectively
\(V(\mathcal{E}_{kj})\)) the surface of the ellipse
\(\mathcal{E}_{ij}\) ( resp . \(\mathcal{E}_{kj}\))
described by the matrix \(P_{ij}\)  (resp. \(P_{kj}\)) of
the Kalman filter of the radar \(i\)  (resp. \(k\)) for the target
\(j\)  and \(V(\mathcal{E}_{ij} \cap \mathcal{E}_{kj})\) 
the intersection volume of these two ellipses.%, as represented on
% \cref{fig-reconciliation-ellipse}:
% \end{itemize}

% $$c_{ikj} = f(V(\mathcal{E}_{ij})) + \alpha g(V(\mathcal{E}_{ij} \cap \mathcal{E}_{kj}))$$

% \begin{figure}[!htb]
% \centering
% \includegraphics[width=.45\textwidth]{media/figure2.pdf}
% \caption{Reconciliation of uncertainty ellipses for two radars following the same target}
% \label{fig-reconciliation-ellipse}
% \end{figure}

% $$\left\{ \begin{matrix}
% f(V(\mathcal{E}_{ij})) > > \alpha g(V(\mathcal{E}_{ij} \cap \mathcal{E}_{kj})) \\
% \alpha g\left( V\left( \mathcal{E}_{ij} \cap \mathcal{E}_{kj} \right) \right) > \varepsilon_{\min},\varepsilon_{\min} \in \mathds{R}_{*}^{+},\text{si }\mathcal{E}_{ij} \cap \mathcal{E}_{kj} \neq \varnothing \\
% \end{matrix} \right.$$

% where:

% \begin{itemize}
\item
\({x_{M}}_{{ij}}\) : Boolean variable, \({x_{M}}_{{ij}}\) 
equals \(1\) if the radar \(i\)  performs the task \(j\)  as the main
radar, 0 otherwise.
\item
\({x_{O}}_{{ij}}\) : Boolean variable, \({x_{O}}_{{ij}}\) 
equals \(1\) if the radar \(i\)  performs the task \(j\)  as an optional
radar, 0 otherwise.
\item
\(w_{{ikj}}\) : Boolean variable, \(w_{{ikj}}\)  equals
\(1\) if the radar \(i\)  performs the task \(j\)  as main radar and the
radar \(k\)  performs the task \(j\)  as optional radar, 0 otherwise.
\end{itemize}

The constraints can be understood the following way:
\begin{itemize}
\item ($A_{ikj}$) defines $w_{ikj}$ as $i$ following the target $j$ as main radar (we also write ${x_{M}}_{{ij}}=1$), and $k$ follows it as optional radar ${x_{O}}_{{kj}}=1$. $w_{iik}=1$ if there is only one radar $i$ following the target. The operator $\wedge$ corresponds to the logical \verb!AND! operator.% The constraint summarized here is equivalent to the two constraints below:
% \[\begin{matrix}
% & {x_{M}}_{{ij}} + {x_{O}}_{{ij}} - 2w_{{ikj}} \leq 1 \\
% & {x_{M}}_{{ij}} + {x_{O}}_{{ij}} - 2w_{{ikj}} \geq 0 \\
% \end{matrix}
% \]

% It may be interesting to note that if $w_{iij} = 1$ then, it
% is considered that the task is only performed by a
% single main radar.
\item
\((C2)\) lists all possible combinations of 2 sensors that
track a target \(j\). There is at most only one combination of sensors
that can be chosen.
\item
\((L)\) models the load of the radar. If the radar is tracking the target as
main or optional radar, the term between parentheses equals 1,
otherwise it equals 0 and the load for the task is therefore not
considered.
\end{itemize}

There is a total of
\(|\mathcal{I}|^{2}\cdot |\mathcal{J}| + |\mathcal{J}| + |\mathcal{I}|\) constraints.

Note that the present formulation is difficult to generalize to a
set of \(n\) sensors, because it would increase the number of constraints and Boolean
variables far too much. Moreover, this would make the problem
insoluble\footnote{Insoluble in a reasonable time. Indeed, the presence
of \(n\) Boolean variables requires performing an enumeration, ie
\(2^{n}\) of possibilities.} for a classical solver. In this paper, we
limit ourselves to 2 radars for each target.

\hypertarget{adaptation-of-distributed-auctions-algorithms-for-multi-radar-tracking}{%
\section{Auction-based multi-radar multi-target allocation}\label{adaptation-of-distributed-auctions-algorithms-for-multi-radar-tracking}}

The approach that we present in this article is based on the use of two
successive CBBA algorithms: the first to make an allocation as the main
radar, the second as an optional radar. In this section, we proceed in
two steps. We first present the CBBA algorithm and the adaptations we
have made to it so that it can consider the specificities of radars. We
then present the general process
allows to consider the interactions between radars and the dynamism of
the mission.

\subsection{The CBBA algorithm}
The Consensus Based Bundle Algorithm \cite{choi2009consensus} is an algorithm where agents bid on a sequence of tasks based on the information they have, and share information with their neighbours. The algorithm can be divided in two phases, that are repeated one after another: (1) the bidding phase during which the agent propose a bid on a sequence (or bundle) of action while trying to optimize the improvement in terms of utility in comparison with the information they have on the current situation. (2) the consensus phase, where the agent send this information, and take information from neighbors, and modify their bundle according to the newly obtained information.

The messages that agents send to each other can be represented as a set of
vectors. The set of vectors that an agent sends to another one
corresponds to its current knowledge of the system. It includes:

\begin{itemize}
\item
\(Y\) the winning bid utility for each target. For a radar \(i\) ,
\(Y = \left( y_{{ij}} \right)_{j \leq |T|}\) 
\item
\(Z\) the identity of the winner for each target. For a radar \(i\) ,
\(Z = \left( z_{{ij}} \right)_{j \leq |T|}\).
\item
\(S\), which corresponds to a ``timestamp'' vector, it makes it
possible to manage conflicts by making it possible to keep the track
of the contacts between radars. For a radar \(i\), \(S_{i} = \left( s_{{ik}} \right)_{k \leq |A|}\). It allows to select the most up-to-date information when there is a conflict among received information.
\end{itemize}

The CBBA algorithm has a 50\% performance guarantee, \emph{i.e.,} in the
worst case, the global solution obtained is greater than half of the
optimal solution. %Certain assumptions are necessary to ensure this
% convergence, the main one being the Diminishing Marginal Gain (DMG)
% constraint. We note \(b_{i}\)  all the tasks allocated to the radar
% \(i\). The DMG constraint is as follows:
% \begin{equation*}
%    c_{{ij}} = c_{{ij}}\left( b_{i} \right) \geq c_{{ij}}\left( b_{i} \oplus b \right) \quad\forall b 
%     \tag{DMG}\label{eq:DMG} 
% \end{equation*}
% The constraint (DMG) reflects the fact that the utility for the same
% target must be decreasing in the number of tasks.
This algorithm is mainly aimed at cases where planning makes sense, for instance when agents are mobile robots and where they must plan a route. But in our case, we must adapt this algorithm in order to make bids on a set of targets, not taking into account the order of the sequence.

\hypertarget{adaptation-of-cbba-to-radars}{%
\subsection{Adaptation of CBBA to radars}\label{adaptation-of-cbba-to-radars}}

In our case, \emph{for the allocation as main radar}, an additional vector
will also be sent, the vector
\(E = \left( e_{{ij}} \right)_{j \leq |T|}\) that groups the
ellipses leading to the winning bids for each target. This notably makes
it possible to calculate intersections with the latter.

In order to respect the Dismininshing Marginal Gain (DMG) constraint that is
required by CBBA, it is necessary to lower the utility of any new target being
tracked. To take acount of this constraint, we introduced the following bias:

\[c_{{ij}}^{\text{CBBA}} = \frac{c_{{ij}}}{|b_{i}|}\]

% \hypertarget{general-loop-interaction-and-dynamism}{%
% \subsection{General loop, interaction and
% dynamism}\label{general-loop-interaction-and-dynamism}}

The algorithm operates in a closed loop and is executed
at each time step; the agent makes an allocation as the
main radar, then it makes the allocation as an optional radar if it has
remaining budget. Each allocation is made through a CBBA algorithm, and
therefore includes the two phases of the algorithm (auction and
consensus) explained above. It therefore receives and sends information
on its allocation as main and optional radar at each time step.% The
% dynamism aspect, which completes the global loop of the algorithm, is
% explained below.
% The interactive aspect can be understood in the following way: on the
% one hand, 
A radar does not take into consideration the targets which it
follows as main radar in the list of targets which it can take as
optional radar.

To implement the interaction between radars, the ellipses
sent by the radars are taken into account by the other radars
to perform the utility calculation for the
allocation as a optional radar. In order to follow as many targets
as possible, when the agent
computes its allocation as main radar, it considers its
budget as all of its remaining budget plus the budget allocated as
optional radar. If a new allocation as main radar is possible, it
deallocates the tasks as optional radar with the lowest utility, and
performs a reset as described in the previous section.

In the static case, when all the radars have the same beliefs on the allocation, we say
that the consensus is established. That is to say that a distributed
allocation conflict-free could be found; this therefore constitutes the
end of the algorithm.

To take account of the dynamic aspect of our problem, the auctions
never stops, and keep running until the end of the simulation.

\begin{enumerate}
\item
The radar makes the bidding phase as the main radar. To do so, it
computes the uncertainty ellipses for each target, and its utility
function; it also applies the Kalman filters of all the targets it is
already tracking.
\item
If it has remaining radar time budget, the radar proceeds to the
bidding phase as an optional radar on all the targets that are not
already tracked as the main radar (with its remaining budget).
\item
The radar proceeds to the consensus phase as the primary radar. The
vectors \(Y,Z,E\) and \(S\) as main radars are updated; vectors are
sent to neighbors.
\item
The radar then proceeds to the consensus phase as an optional radar.
Vectors \(Y,Z\) and \(S\)  as optional radars are updated; vectors are
sent to neighbors.
\item
The radar tracks the targets it has selected, possibly by applying its
Kalman filter.
\end{enumerate}

Note that since the radar initiates the tracks at each execution of the
two phases of CBBA, which means that the algorithm does not have time to
converge. In practice, this situation can generate conflicts, for instance in the case
where a target on the edge between two radars, and is increasingly
threatening. In this case, each radar takes its decision on a previous
value of the utility of the others, and its own current value, it 
considers that its bid wins the bet. Similarly, targets getting
less threatening (\emph{i.e.,} with decreasing utility) is
potentially tracked by none of the radars, each radar considering that another
one has a better bid than itself.

\begin{figure}[!tp]
	\centering
	
	\begin{subfigure}{.3\textwidth}
	\centering
		\includegraphics[width=\textwidth]{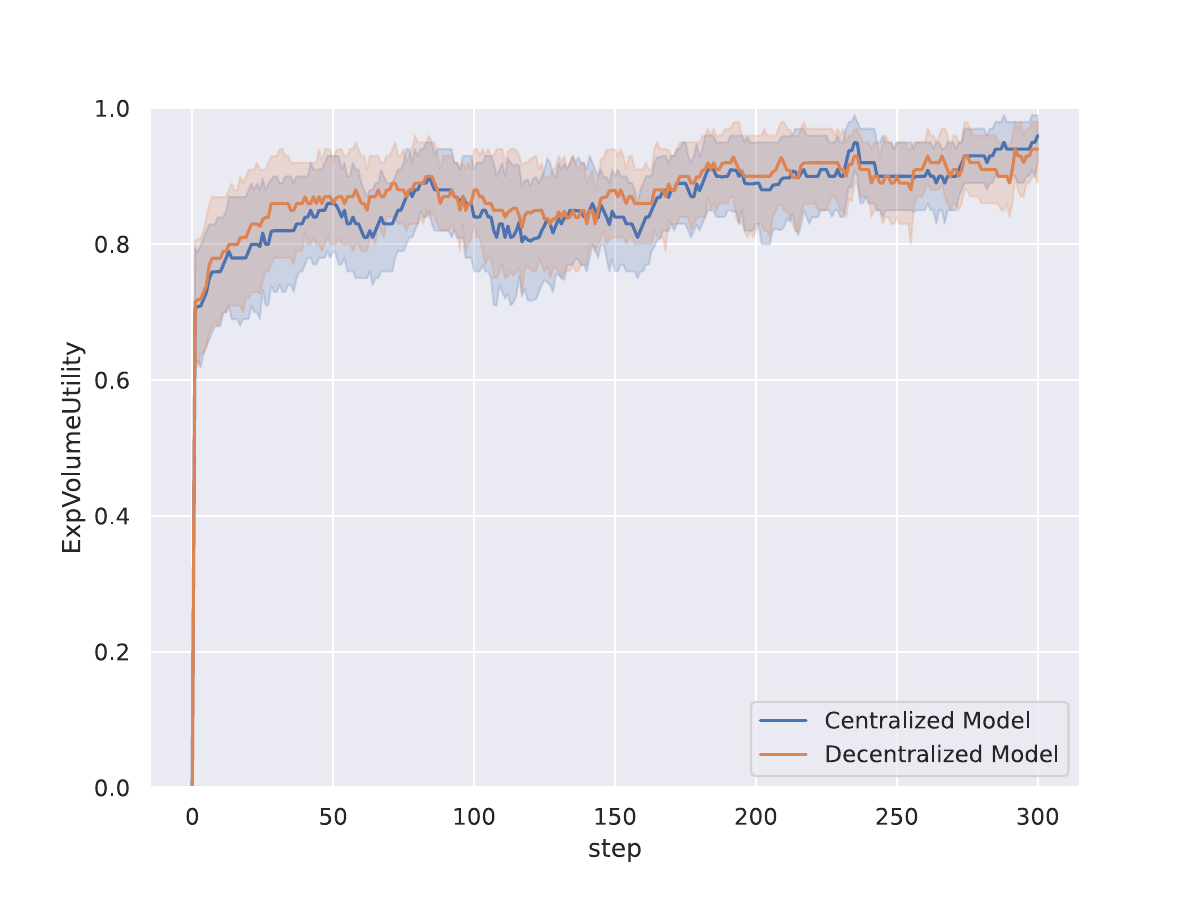}
		\caption{Non saturated well positioned radars}
	\end{subfigure}
	\begin{subfigure}{.3\textwidth}
	\centering
		\includegraphics[width=\textwidth]{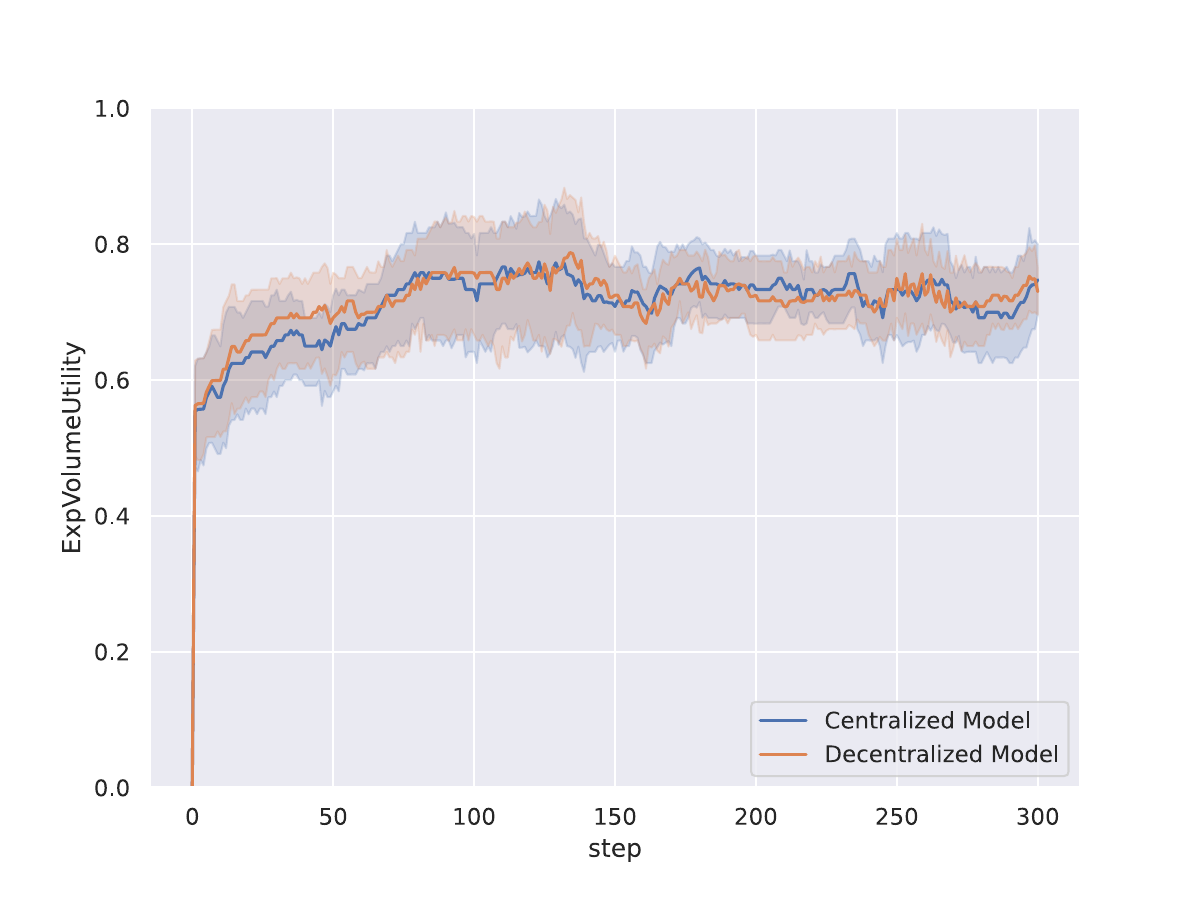}
		\caption{Few saturated well positioned radars}
	\end{subfigure}
	\begin{subfigure}{.3\textwidth}
		\centering
		\includegraphics[width=\textwidth]{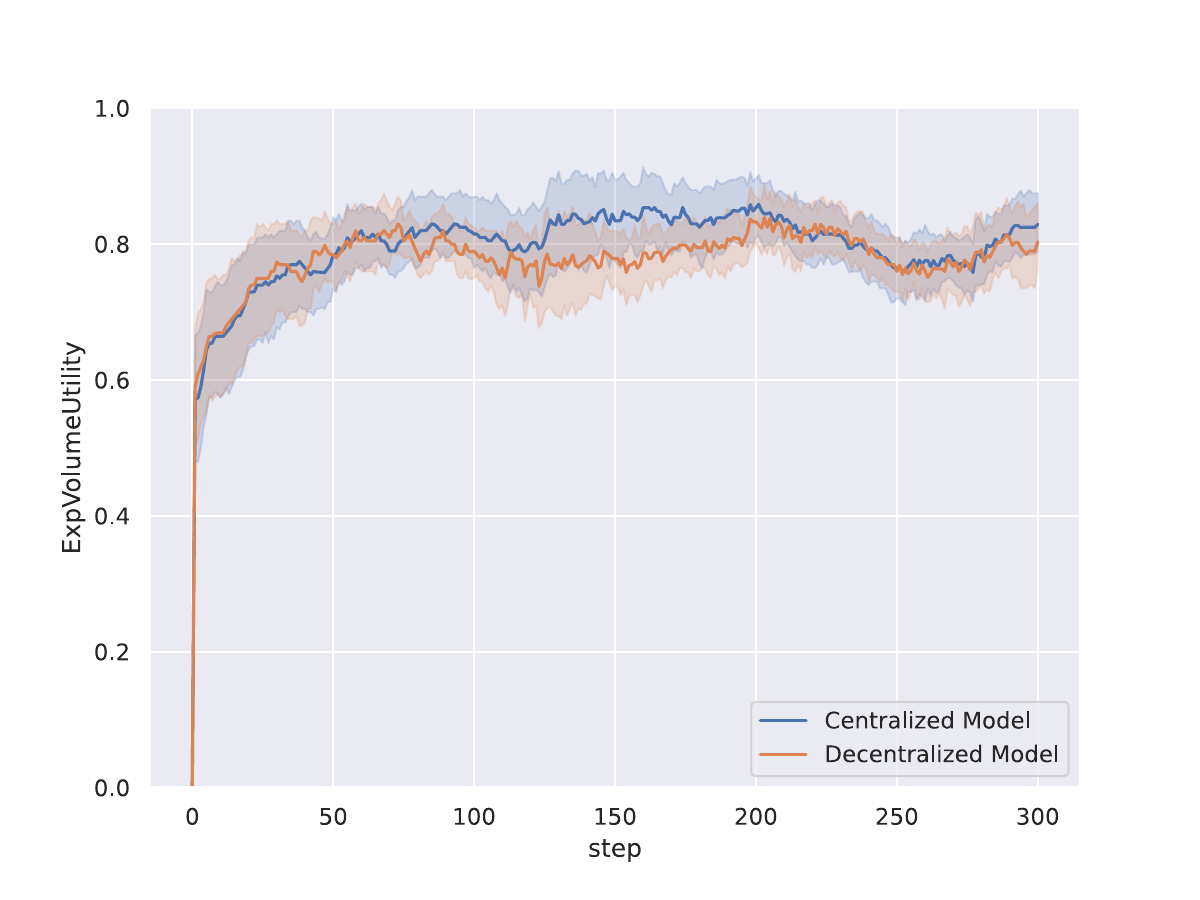}
		\caption{Several saturated well positioned radars}
	\end{subfigure}
% \end{figure}

% \begin{figure}
% \ContinuedFloat
% \centering
	\begin{subfigure}{.4\textwidth}
		\centering
		\includegraphics[width=\textwidth]{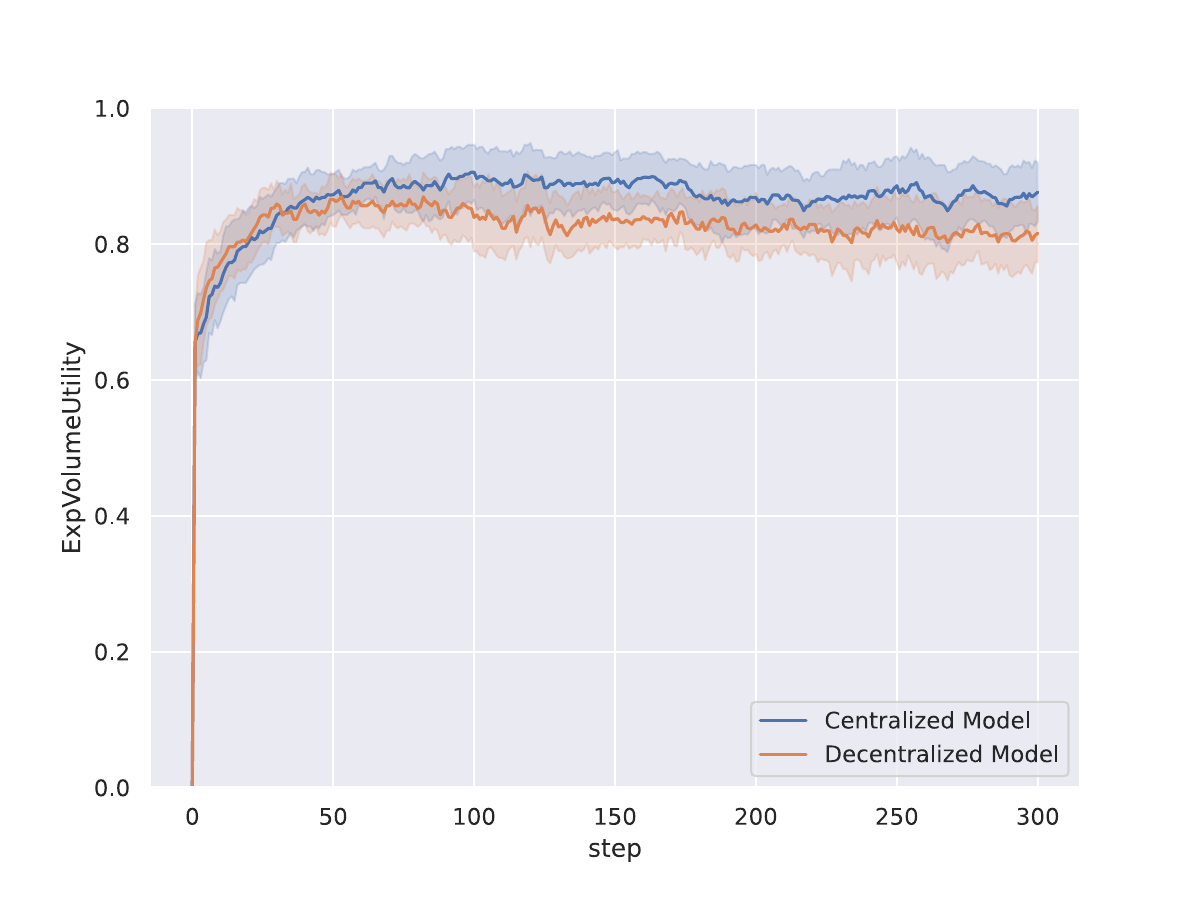}
		\caption{Many saturated well positioned radars}
	\end{subfigure}
\begin{subfigure}{.4\textwidth}
		\centering
		\includegraphics[width=\textwidth]{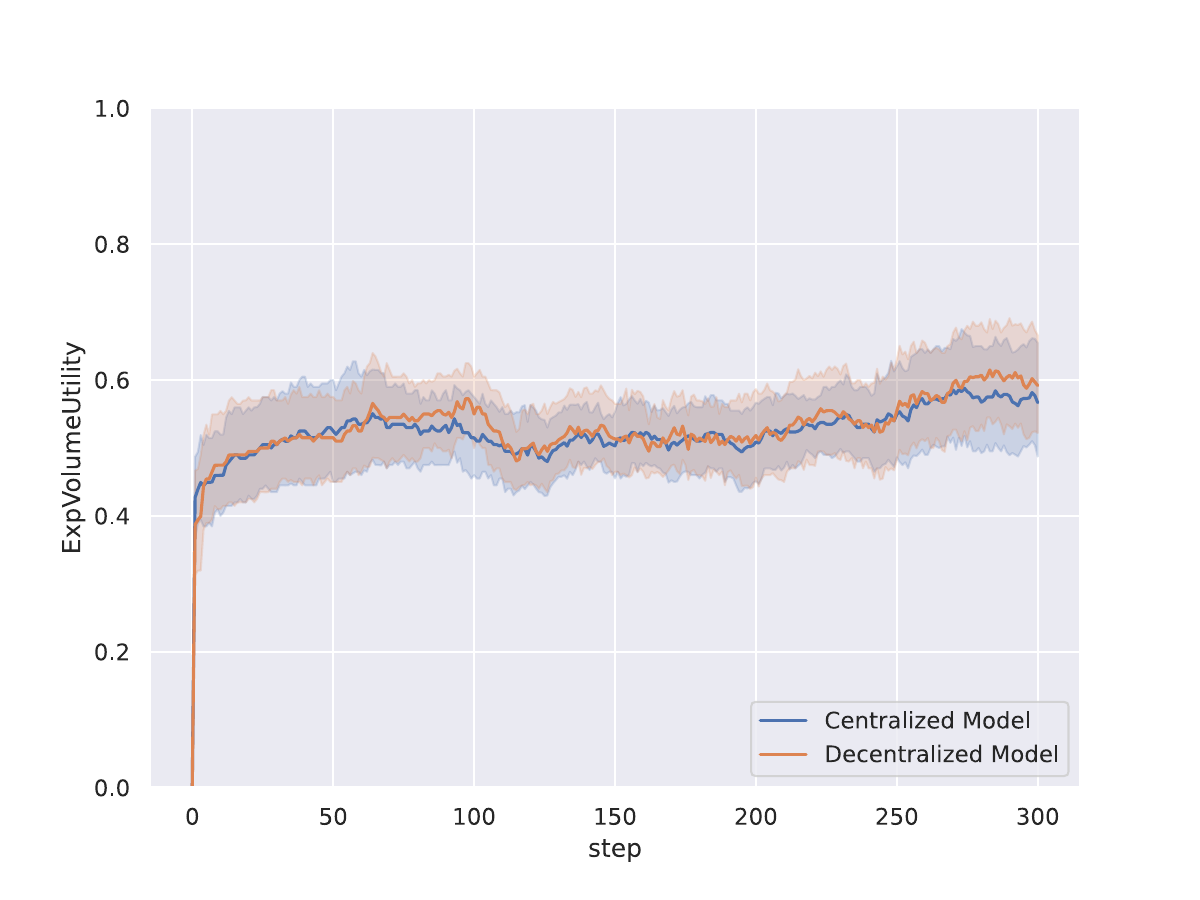}
		\caption{Saturated ill-positioned radars}
	\end{subfigure}
\caption{Utility on the reference scenarios}
\label{fig-utility}
\end{figure}

\begin{figure}[!tp]
	\centering
	
	\begin{subfigure}{.3\textwidth}
	\centering
		\includegraphics[width=\textwidth]{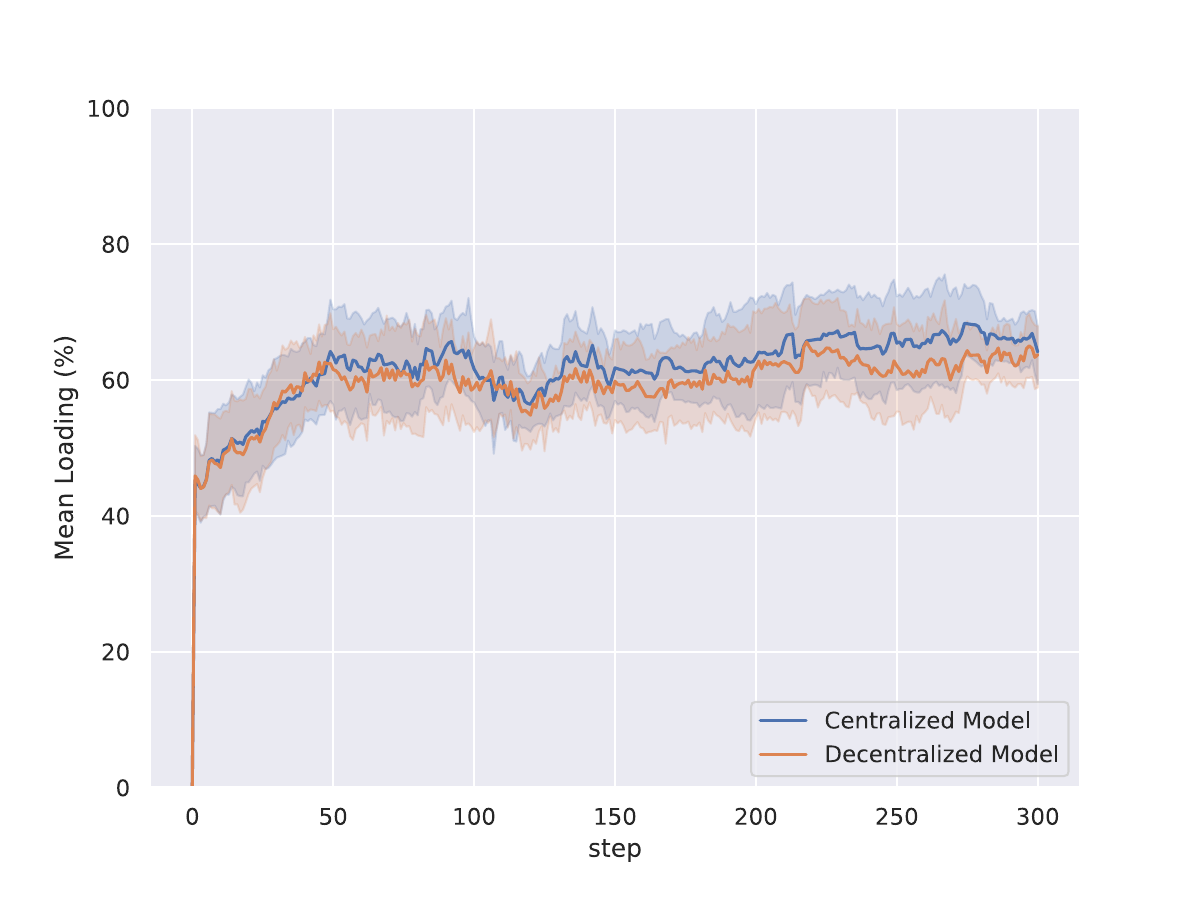}
		\caption{Non saturated well positioned radars}
	\end{subfigure}
	\begin{subfigure}{.3\textwidth}
	\centering
		\includegraphics[width=\textwidth]{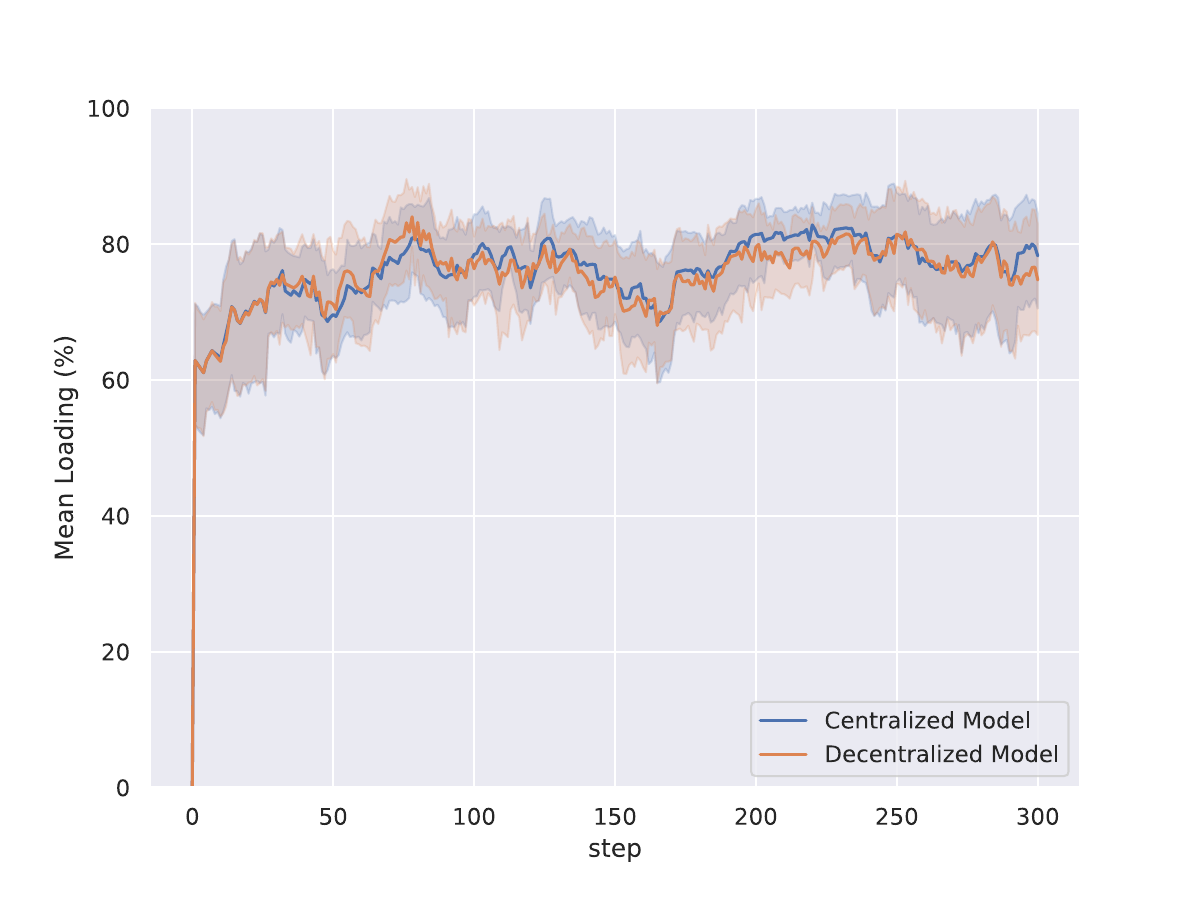}
		\caption{Few saturated well positioned radars}
	\end{subfigure}
	\begin{subfigure}{.3\textwidth}
		\centering
		\includegraphics[width=\textwidth]{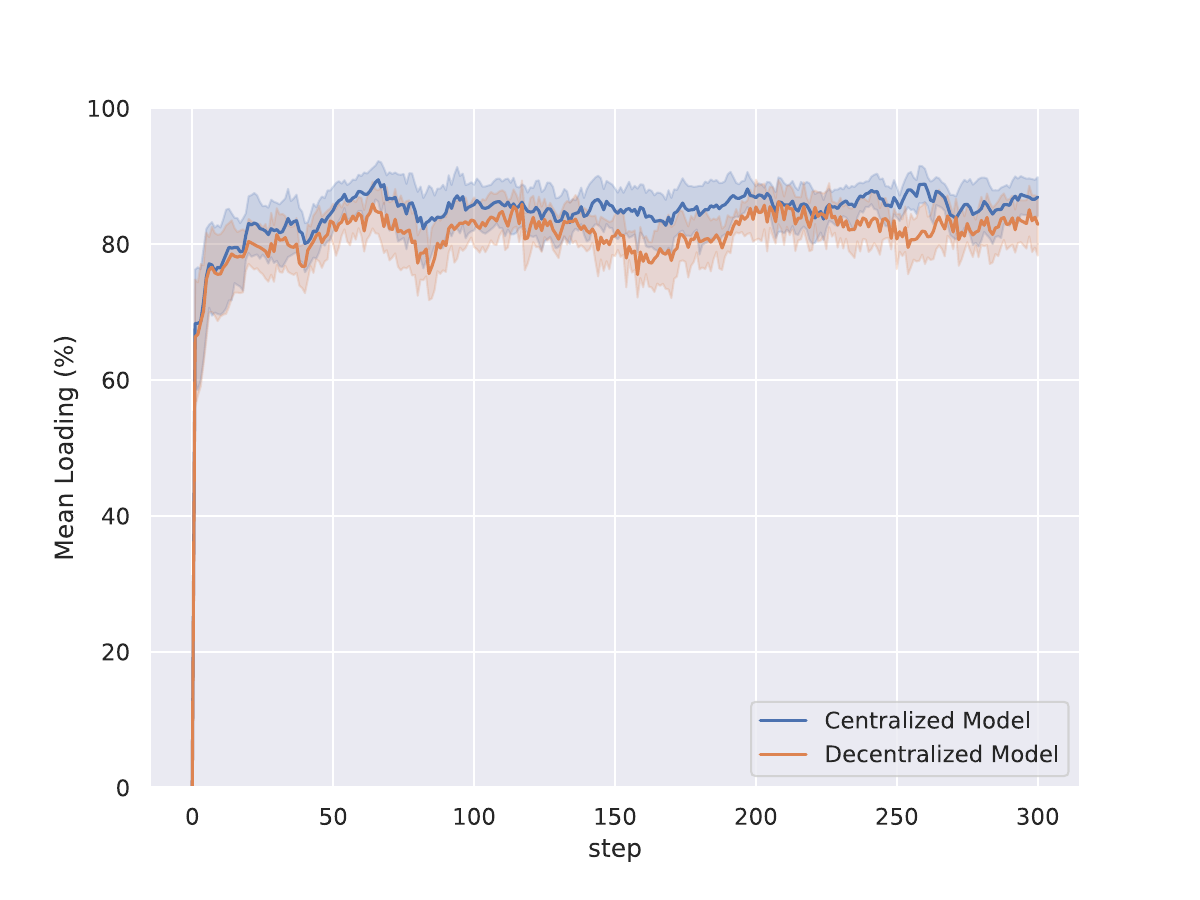}
		\caption{Several saturated well positioned radars}
	\end{subfigure}
% \end{figure}

% \begin{figure}
% \ContinuedFloat	
% \centering
	\begin{subfigure}{.4\textwidth}
		\centering
		\includegraphics[width=\textwidth]{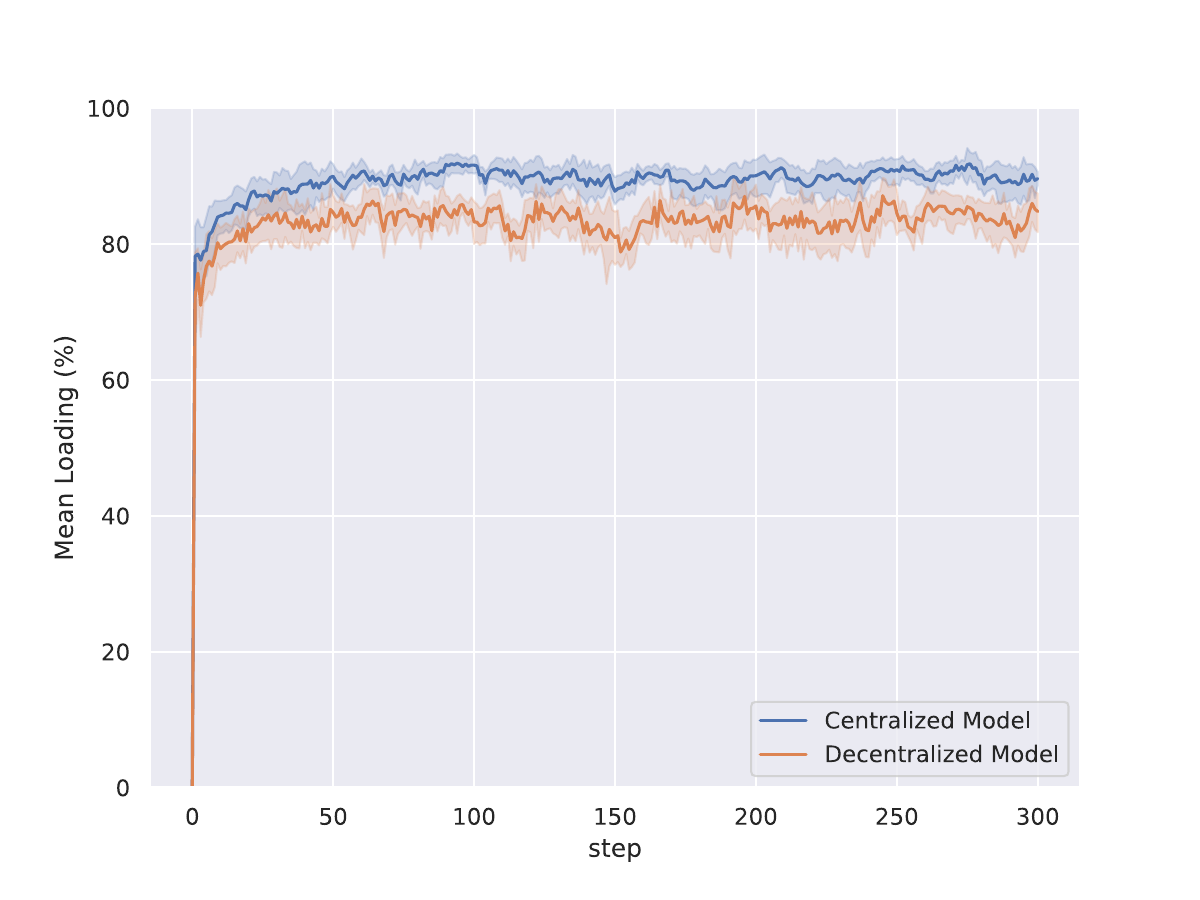}
		\caption{Many saturated well positioned radars}
	\end{subfigure}
	\begin{subfigure}{.4\textwidth}
		\centering
		\includegraphics[width=\textwidth]{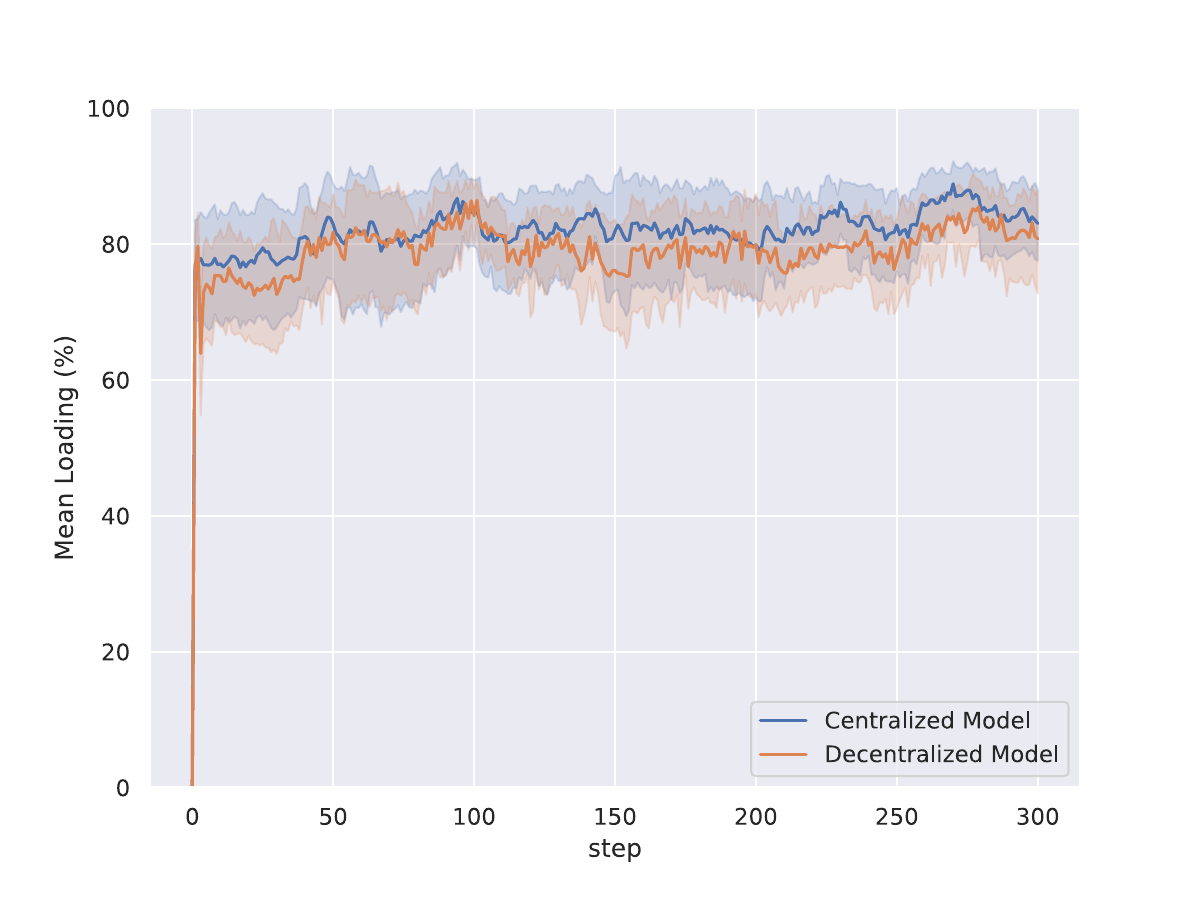}
		\caption{Saturated ill-positioned radars}
	\end{subfigure}
\caption{Load on the reference scenarios}
\label{fig-load}
\end{figure}

% \begin{figure}
% 	\centering
% 	\begin{subfigure}{.24\textwidth}
% 		\includegraphics[width=\textwidth]{media/image3.png}
% 		\caption{initial allocation}
% 	\end{subfigure}
% 	\begin{subfigure}{.24\textwidth}
% 		\includegraphics[width=\textwidth]{media/image4.png}
% 		\caption{sending messages}
% 	\end{subfigure}
% 	\begin{subfigure}{.24\textwidth}
% 		\includegraphics[width=\textwidth]{media/image5.png}
% 		\caption{Update (consensus)}
% 	\end{subfigure}
% 	\begin{subfigure}{.24\textwidth}
% 		\includegraphics[width=\textwidth]{media/image6.png}
% 		\caption{Bidding \& sending messages}
% 	\end{subfigure}
% 	\begin{subfigure}{.24\textwidth}
% 		\includegraphics[width=\textwidth]{media/image10_2.png}
% 		\caption{Second consensus phase, consensus reached}
% 	\end{subfigure}
% \caption{Master Allocation Method Workflow for Two Targets and Three Radars (with CBAA)}
% \end{figure}

% \hypertarget{algorithm-illustration}{%
% \section{Algorithm illustration}\label{algorithm-illustration}}

Our approach makes it possible to solve the task allocation problem in
an advantageous way: in the case where the radars remain in contact,
even if the communication graph evolves during the mission, the algorithm
keeps working as far as graph of connection of the radars remains
path-connected. Finally, by
carrying out two sequential auction runs, our algorithm also takes into
account the possible overlapping of uncertainty ellipses, and thus makes
it possible to generate an allocation favoring more precise tracking of
targets when possible, while trying to track as many
targets as possible (depending on radar capability).

Radars must also be able to differentiate between first-round allocation
messages -- one that tracks as many targets as possible -- and one that
improves accuracy by generating an intersection of uncertainty ellipses.
Each radar \(i\)  proceeds by trying to maximize
\(\sum_{j}{x_{{ij}} \cdot}c_{{ij}}\) for the main allocation
(resp.\(\sum_{j,k}{w_{{ikj}} \cdot}c_{{ikj}}\)  for the
optional allocation), that is to say the sum of the utilities
corresponding to the targets that it tracks, while taking into account
the information received from the other radars, in particular the bids
made by the latter.

% An example of the way the algorithm works for allocation of the main
% radar is represented on \cref{simu-radars}.
The implementation must include
an additional target disambiguation mechanism, making it possible to
identify the targets present at several radars, and in particular a plot
merging algorithm, making it possible to match the targets of the
different radars. This induces the sending of additional information
enabling this operation to be carried out, such as the estimated speed
and position of the targets.

% \begin{figure}[!htb]
% 	\centering
% 	\includegraphics[width=.29\textwidth]{media/image8.png}
% 	\caption{Mesa simulator}
% 	\label{simu-radars}
% \end{figure}

\hypertarget{results}{%
\section{Results}\label{results}}

The implementation of our model has been performed on the MESA framework
\cite{kazil2020utilizing}, along with the Kalman filter package \cite{laaraiedh2012implementation},
on the same simulator as the one used in \cite{nour2021multi}. As in this article, we rely on 5 kinds of scenarios, similar bu not identical to those of \cite{nour2021multi}; results are averaged on 10 such scenarios for each graph:
\begin{itemize}
    \item Non saturated well positioned radars (5 radars, 10 targets)
    \item Few saturated well positioned radars (3 radars, 12 targets)
    \item Several saturated well positioned radars (5 radars, 20 targets)
    \item Many saturated well positioned radars (8 radars, 30 targets)
    \item Saturated ill-positioned radars (4 radars, 20 targets)
\end{itemize}

% The \cref{simu-radars}
% shows the simulator: the radars are represented as blue points while the
% targets are represented as red triangles. Main radars following a target
% are represented as green lines, optional ones as purple lines. The
% uncertainty ellipses are in yellow.

In order to evaluate our work, we compare it to an centralized allocation,
performed with the Coin-OR Branch and Cut tool (CBC) \cite{forrest2005cbc}. Results are
represented on \cref{fig-utility,fig-load} and are averaged on 10 similar scenarios. The blue (resp. orange) curves correspond to the
decentralized (resp. centralized) approach. The standard deviation is represented in light blue/light orange.

The utility of the system on the scenarios is represented on \cref{fig-utility}. The decentralized approach is pretty close to the results of the centralized approach, in particular for situations where the radars are not saturated. This is pretty interesting because for similar utility, the decentralized approach gets lower load. This can be explained by the fact that the optimization is performed only on the current step and the centralized approach computes a new allocation through optimization algorithm each step, while the CBBA only makes changes if it has remaining budget, which makes it more ``stable'', and therefore improves the result of the Kalman filters. It is also the case for ill-positioned radars, where the ``jumps'' between radars do not have big influence on the utility, as the improvement is equivalent to the loss in terms of the Kalman filter. The case with few radars is almost similar for the centralized and decentralized approaches, the gaps and loss in optimality balancing each other.

In cases where the radars are saturated and numerous, on the contrary the radars jump from target to target for both approaches but centralized approach makes a more complete exploration of the possible allocations. This explains the higher utility for the centralized approach and also the higher load. The coverage (proportion of tracked targets, not presented for space reasons) is also higher for the centralized approach, which confirms our hypothesis.

Overall, our approach performs almost as well as the centralized approach. When the radars are not saturated, it even performs better. In cases where the radars are saturated and numerous, a less complete research leads to targets not being tracked, and the centralized approach is better.

\hypertarget{conclusion}{%
\section{Conclusion}\label{conclusion}}

In this paper, we presented a novel approach for allocating target to a
team of radars in a totally decentralized way. This approach is based on
a fully decentralized auction algorithm, CBBA. We showed that, when
taking into account the intersection of uncertainty ellipses, the
results of this algorithm is comparable to the centralized
allocation, better in cases where the radars are not too numerous and not saturated, and a bit worse when the radars are numerous and saturated.

Future works include the design of a more generic approach handle an arbitrary number of radars following the same target. We also
would like to make our approach more dynamic, for instance by including
replanning approaches that have been proposed to improve CBBA
\cite{buckman2019partial} and evaluate this approach in the setting imposed by our use-case. Finally, including a possible variation of the SNR would allow a finer allocation and therefore improve our approach.

% \newpage
\bibliographystyle{unsrt}
\bibliography{biblio}
\end{document}